\documentclass{cta-author}

{}
{}
{}

\usepackage{times}
\usepackage{epsfig}
\usepackage{graphicx}
\usepackage{amsmath}
\usepackage{amssymb}
\usepackage{caption}
\usepackage{subcaption}
\usepackage{float}
\usepackage[normalem]{ulem}
\usepackage{multirow}
\usepackage{arydshln}
\usepackage{xcolor}
\usepackage{siunitx}
\usepackage{hyperref}

\useunder{\uline}{\ul}{}

\begin{document}

\title{300 GHz Radar Object Recognition based on Deep Neural Networks and Transfer Learning}

\author{\au{Marcel Sheeny $^{1}$}
\au{Andrew Wallace $^1$}
\au{Sen Wang $^1$}
}

\address{\add{1}{Institute of Sensors, Signals and Systems, Heriot-Watt University, Edinburgh, EH14 4AS, UK}
\email{ms47@hw.ac.uk, a.m.wallace@hw.ac.uk,s.wang@hw.ac.uk}}

\begin{abstract}

For high resolution scene mapping and object recognition, optical technologies such as cameras and LiDAR are the sensors of choice. However, for robust future vehicle autonomy and driver assistance in adverse weather conditions, improvements in automotive radar technology, and the development of algorithms and machine learning for robust mapping and recognition are essential. In this paper, we describe a methodology based on deep neural networks to recognise objects in 300GHz radar images, investigating robustness to changes in range, orientation and different receivers in a laboratory environment. As the training data is limited, we have also investigated the effects of transfer learning. As a necessary first step before road trials, we have also considered detection and classification in multiple object scenes.

\end{abstract}

\maketitle

\section{Introduction}

All major car manufacturers are evaluating LiDAR, passive optical and radar sensing capabilities for automotive applications \cite{IbaZeaCon2018}, aiming beyond advanced driver-assistance systems (ADAS) such as automatic cruise control, parking assistance and collision avoidance, towards full automotive autonomy. 
Each technology has benefits and drawbacks, but a key benefit of automotive radar is an operating range up to $150m$ or more, and an ability to function in adverse weather, such as fog, rain or mist.
However, radar sensors offer much lower resolution than optical technologies.
Current automotive radar systems operate at 24 GHz and 79 GHz, with a typical bandwidth of 4 GHz, are able to perform low resolution mapping and detection in relatively uncluttered scenes, but object recognition is really challenging.

Deep Neural Networks (DNNs) have proven to be a powerful technique for image recognition on natural images \cite{krizhevsky2012imagenet, szegedy2015going, Simonyan15}. 
In contrast to manual selection of suitable features, followed by statistical classification, DNNs optimise the learning process to find a wider range of
patterns, achieving better results than formerly on quite complicated scenarios,  including for example the ImageNet challenge first introduced in 2009 \cite{deng2009imagenet}, which has at the time of writing more than 2000 object categories and 14 million images.



In this paper, we wish to assess the capability of DNNs applied to images of objects acquired by a prospective 300 GHz automotive radar with an operating bandwidth of 20GHz. Our experiments, conducted in a laboratory setting from late 2017  uses a small database of 6 Isolated objects to assess the current capability. Later, in August 2019, we trained our neural networks in a more challenging scenario with multiple objects in the same scene to assess the performance to both detect and classify objects in the presence of both uniform and cluttered background.

The principal contributions of the work are to assess the robustness of the DNNs to variations in viewing angle, range and specific receiver; since we have limited data, we also investigated how transfer learning can improve the results. We also evaluate the performance of the trained neural networks in a more challenging scenario with multiple objects in the same scene.
Using DNNs, we classify these objects with minimum domain knowledge about the sensors and objects being sensed.
The scenes are static; we do not use range-doppler spectra to classify images, but perform experiments on the radar power data alone.

\begin{figure}[!]
\centering
\includegraphics[width=\linewidth]{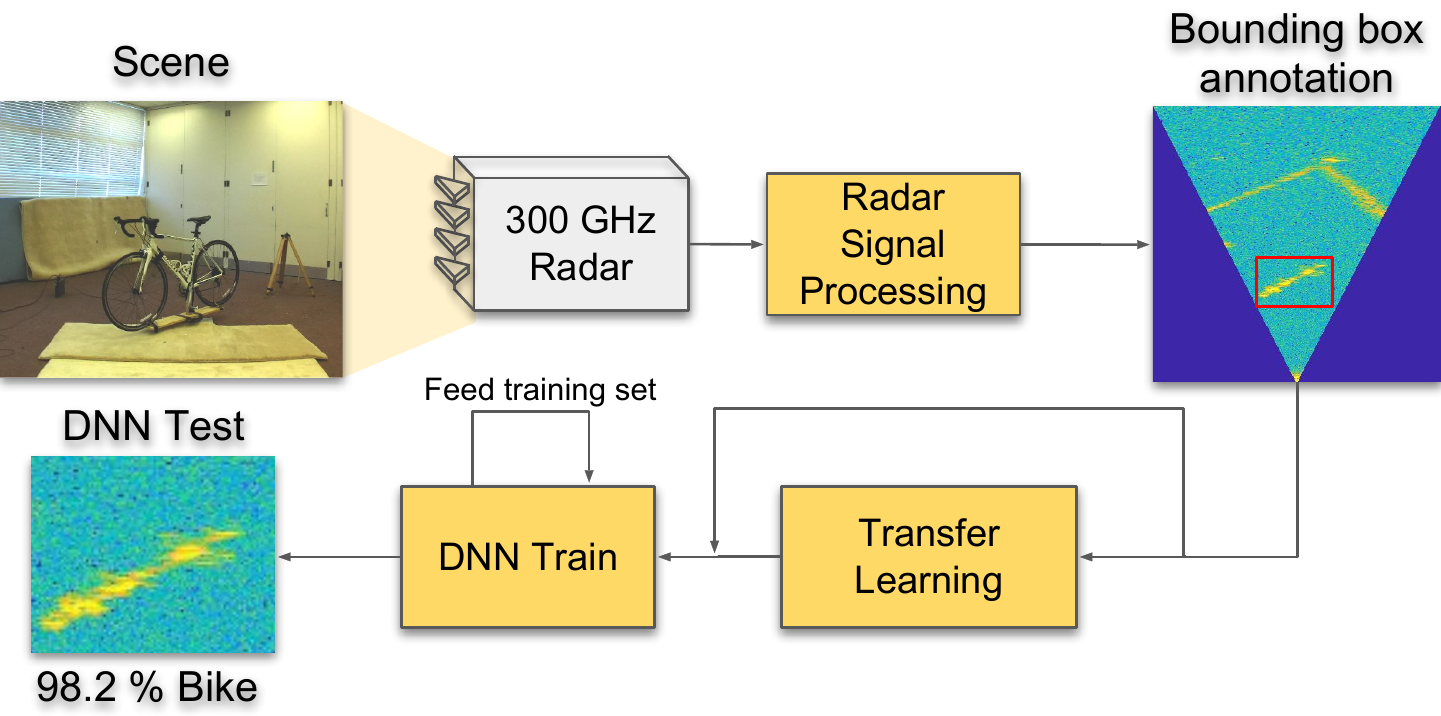}
\caption{\textbf{300 GHz FMCW Radar Object Recognition:} Methodology developed using deep convolutional neural networks to process data acquired by a prototype high resolution 300 GHz short range radar \cite{Daniel2018}. Steps: 1. Radar Signal Processing: Cartesian radar image generation. 2. Bounding box annotation to crop object region. 3. Deep Neural Network and Transfer Learning radar based recognition.}
\label{fig:methodology}
\end{figure}

\section{Related Work}

Together with scene mapping, object recognition is a necessary capability for autonomous cars. When we create a map of the immediate environment, we also need to identify key actors, such as pedestrians and vehicles, and other street furniture, traffic signs, walls, junctions and so on. For actors, we would also wish to predict their movement in order to create a safe system, and identity is  a key component of such prediction.

The use of deep convolutional neural networks (DCNNs) \cite{lecun1995convolutional,krizhevsky2012imagenet} for large scale image recognition has changed significantly the field of computer vision. Although questions remain on verifiability \cite{huang2017safety}, confidence in the results \cite{gal2016dropout}, and on the effects of adversarial examples  \cite{nguyen2015deep}, the best results for correct identifications applied to large image datasets have been dominated by DCNNs algorithms. The development of GPU's and large annotated datasets has helped the popularity of deep learning methods in computer vision.

Of course, the results on natural image data such as ImageNet can be replicated to a large extent using automotive data, such as the KITTI benchmarks \cite{geiger2012we}.
However, in adverse weather, those sensors have poor performance, so we wish to examine the potential of radar data for reliable recognition.
This is especially challenging; most automotive radars sense in two dimensions only, azimuth and range, although research is underway to develop full 3D radar \cite{phi300_2017}.
Although range resolution can be of the orders of $cm$, azimuth resolution is poor, typically $1-2$ degrees although again there is active research to improve this \cite{Daniel2018}.
Natural image recognition relies to a great extent on surface detail, but the radar imaging of surfaces is much less well understood, is variable, and full electromagnetic modelling of complex scenes is extremely difficult.
 

There has been some recent work in applying deep learning techniques to radar images for automotive applications. Wohler \etal \cite{wohler2017comparison} \cite{schumann2018semantic} have used Long-Term Short Memory neural networks creating a methodology to classify road actors in the automotive scenario.
Lombacher \etal \cite{lombacher2017semantic} also used deep learning techniques to segment cars against other objects.
These examples use only power data. Why not use readily available motion data available from Doppler shift?
Rohling \etal \cite{rohling2010pedestrian} used a 24GHz radar to classify pedestrians by analysing the Doppler spectrum and range profile. Similarly Bartsch et al. \cite{bartsch2012pedestrian}  classified pedestrians using the area and shape of the object and Doppler spectrum features. They analysed the probability of each feature and used a simple decision model. They achieved 95\% classification rates for optimal scenarios, but this dropped to 29.4\% when the pedestrian was in close proximity to cars due to low resolution from the radar sensors.
Likewise Angelov \etal \cite{angelov2019} investigated the capability of different DCNNs to recognise cars, people and bicycles with variable success rates ranging from tests accuracies of 44-88\% depending on the problem.
The conclusion from these studies is that prototypical motion can be a powerful aid to object identification, but with powerful caveats.
First, a car is still a car if stationary at traffic lights, and second, for a moving ego-vehicle the whole scene is moving, not just readily separable targets.

\section{Applying Deep Neural Networks to 300 GHz Radar Data}

\subsection{Objective}

The main objective of this work is to design and evaluate a methodology for object classification in 300 GHz radar data using DCNNs, as illustrated schematically in Figure \ref{fig:methodology}. 
This is a prototype radar system; we have limited data so we have employed data augmentation and transfer to examine
whether this improves our recognition success.
To verify the robustness of our approach,
we have assessed recognition rates using different receivers at different positions, and objects at different orientations and range. We also evaluated the performance of the method in a more challenging scenario with multiple objects per scene.

\subsection{300 GHz FMCW Radar}

A current, typical commercial vehicle radar uses  MIMO technology at 77-79GHz with up to 4 GHz IF bandwidth, and a range resolution of 4.3-35cm dependent on target range, 20-80m, and an azimuth resolution of 15 degrees \cite{tiradar}. This equates to a cross range resolution of $\approx 4m$ at $15m$ such that a car will just occupy one cell in the radar image. This is clearly not sensible for object recognition on the basis of radar cross section. In this work, we collected data using a FMCW 300 GHz scanning radar designed at the University of Birmingham \cite{phi300_2017}. The main advantage of the increased resolution is a better radar image which may lead to more reliable object classification. The 300 GHz radar used in this work has a bandwidth of 20 GHz which equates to 0.75 cm range resolution. The azimuth resolution is $1.2^o$ which corresponds to $20cm$ at 10 meters.
The parameters for the 300 GHz sensor used in this work can be seen in Table \ref{tab:300ghz_radar}.

\begin{table}[h!]
\caption{300 GHz FMCW Radar parameters for the system described in \cite{phi300_2017}.} \label{tab:300ghz_radar}
\centering
\begin{tabular}{r|c}
\hline
Sweep Bandwidth               & $20$ GHz                    \\ 
H-Plane (Azimuth) beamwidth   & $1.2^o$                    \\ 
E-Plane (Elevation) beamwidth & $1.6^o$            \\  
Antenna gain                  & $39$ dBi                   \\ 
Range resolution              & $0.75$ cm                     \\ 
Azimuth resolution (at 10 m)  & $20$ cm                    \\ 
Elevation resolution (at 10 m)  & $28$ cm                    \\ 

\hline
\end{tabular}

\end{table}

\begin{figure}[t]
    \centering
    \includegraphics[width=\linewidth]{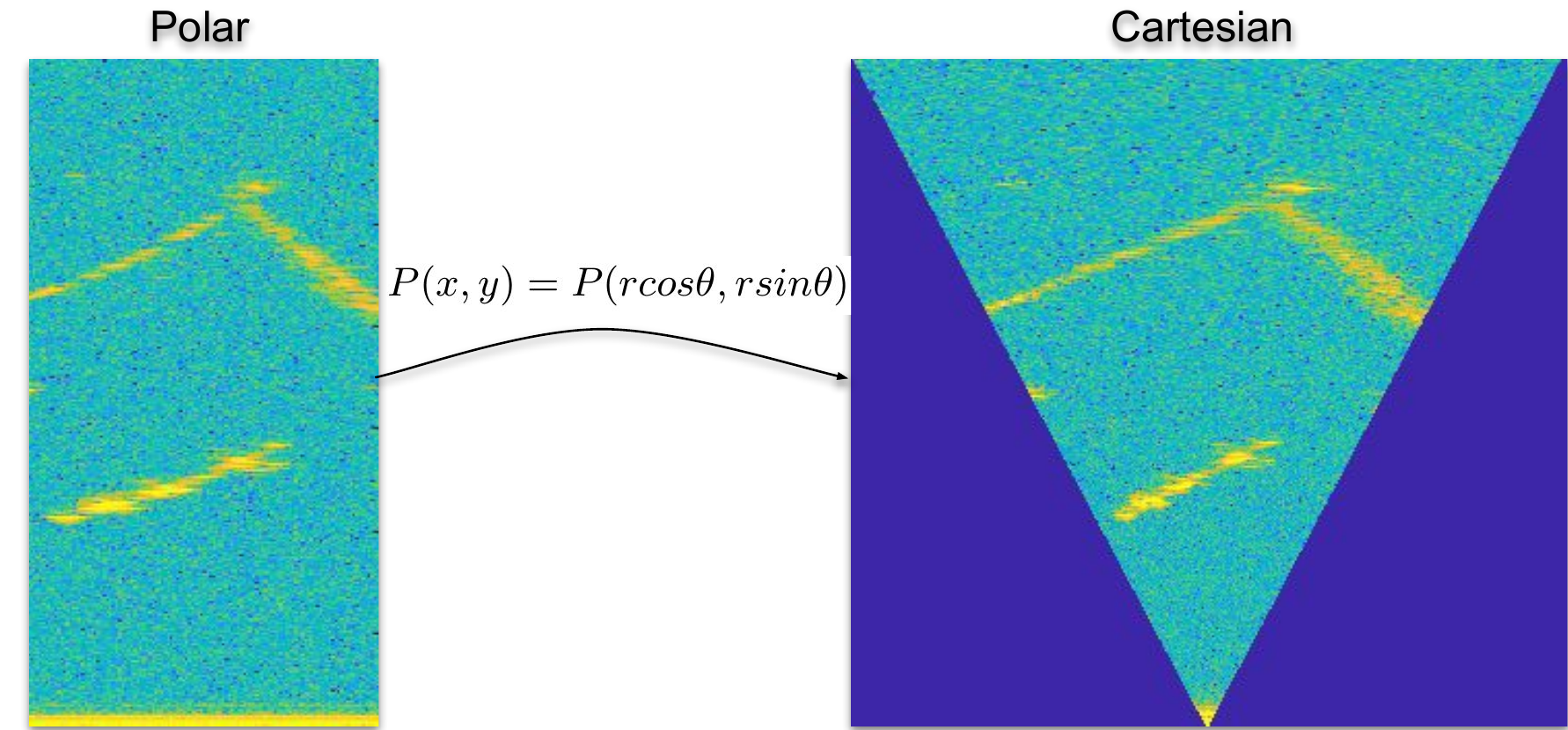}
    \caption{Polar to Cartesian radar image}
    \label{fig:pol2cart}
\end{figure}

The raw data captured by the 300 GHz radar is a time-domain signal at each azimuth direction. To transform the raw signal into an image two steps were performed. The first step is to apply Fast Fourier Transform (FFT) to each azimuth signal to create a range profile. The original polar image is converted to cartesian coordinates as shown in Figure \ref{fig:pol2cart}. 
Before training the neural network with this data, we applied whitening by subtracting the mean value of the image data, as this helps the stochastic gradient descent (SGD) to converge faster.

\subsection{Experimental Design and Data Collection}\label{sec:data_collected}

The main objective is to establish whether the proposed methodology has the potential to discriminate between a limited set of prototypical objects in a laboratory scenario, prior to collecting wild data in a scaled down or alternate radar system.
We wanted to gain knowledge of what features were important in 300 GHz radar data, and whether such features were invariant to the several possible transformations. The objects we decided to use were a bike, trolley, mannequin, sign, stuffed dog and cone. Those objects contain a varieties of shapes and materials which to some extent typify the expected, roadside  radar images that we might acquire from a vehicle.

\begin{figure}[!]
\centering
\includegraphics[width=0.9\linewidth]{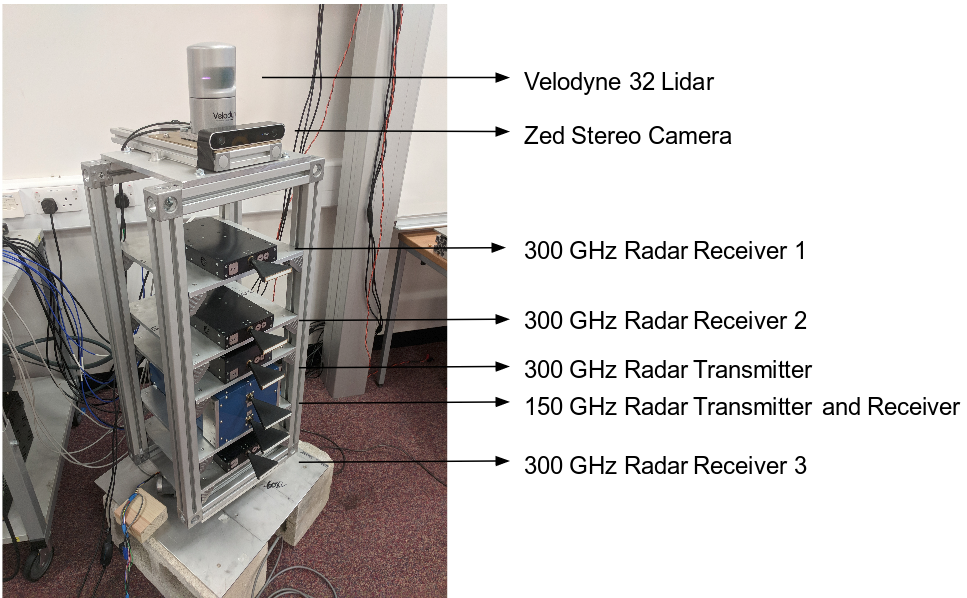}
\caption{Experimental sensor setup}
\label{fig:exp_config}
\end{figure}

\begin{figure}[!]
\centering
\includegraphics[width=0.8\linewidth]{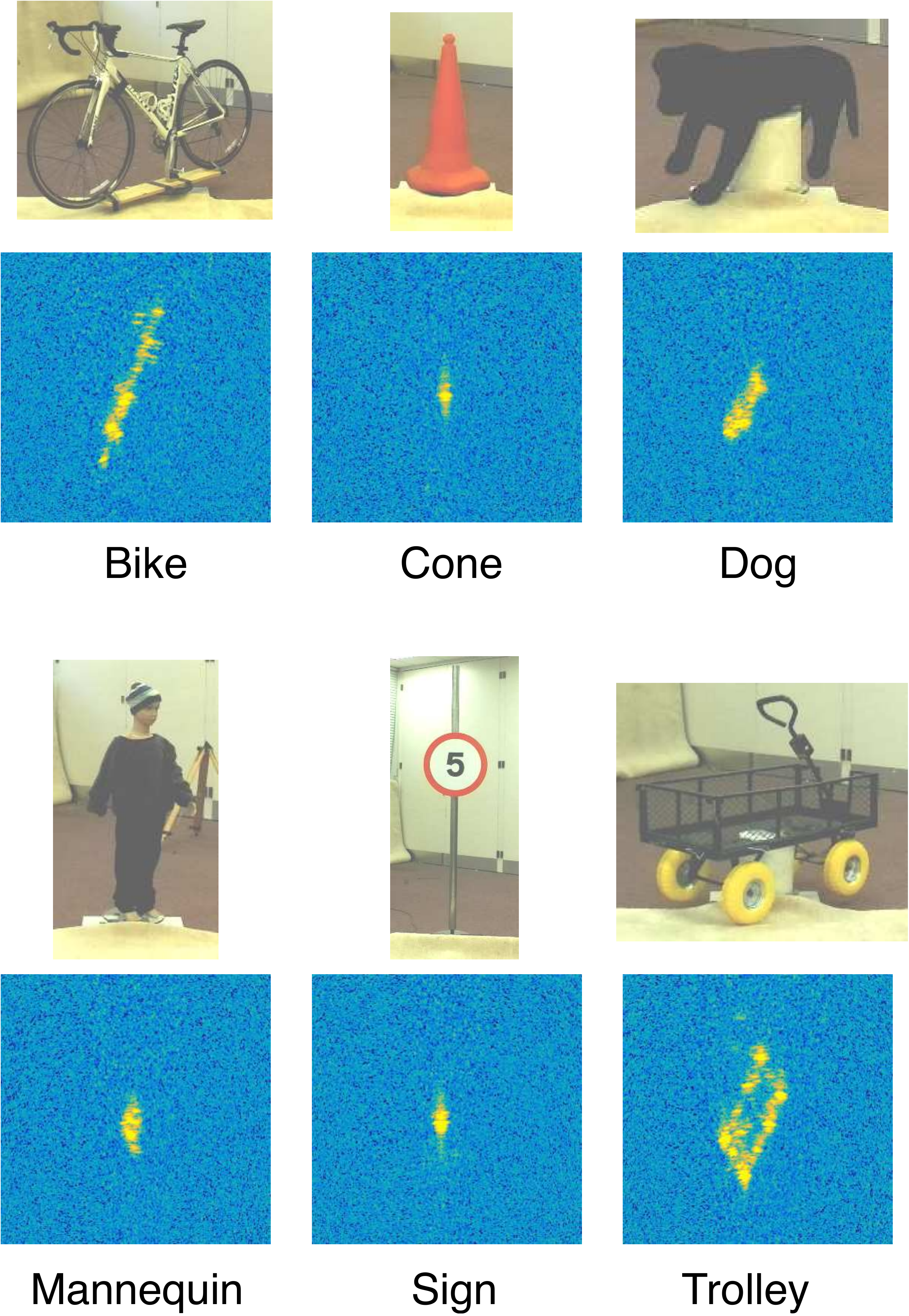}
\caption{Sample images from each object from the dataset collected using the 300 GHz radar}
\label{fig:samples}
\end{figure}

The equipment for automatic data collection included a turntable to acquire samples every 4 degrees, covering all aspect angles, and at two stand-off distances, 3.8 m and 6.3 m. The sensors are shown in Figure \ref{fig:exp_config}. In collecting data, We used 300 GHz and 150 GHz radars, a Stereo Zed camera and a Velodyne HDL-32e Lidar, but in this paper only data from the 300 GHz radar is considered. The 300 GHz radar has 1 transmitter and 3 receivers. The 3 receivers were used to compare the object signatures at different heights. We used a carpet below the objects to avoid multi-path and ground reflections.
Table \ref{tab:tab_dataset} summarises how many samples were captured from each object at each range. Since we have 3 receivers, we have 1425 images from each range and 2850 images in total. In Figure \ref{fig:samples} we can see sample images from all objects at different ranges using receiver 3.

\begin{table}[!]
\caption{Data set collection showing number of different raw images collected at each range.}\label{tab:tab_dataset}
\begin{center}
\begin{tabular}{lcc}
\hline
	& \textbf{3.8 m}	& \textbf{6.3 m}\\
\hline
Bike         & 90                                             & 90                                             \\
Trolley      & 90                                             & 90                                             \\
Mannequin    & 90                                             & 90                                             \\
Cone         & 25                                             & 25                                             \\
Traffic Sign & 90                                             & 90                                             \\
Stuffed Dog  & 90                                             & 90                                             \\ \hline
Total        & \multicolumn{1}{l}{475 $\times$ (3 rec.) = 1425} & \multicolumn{1}{l}{475 $\times$ (3 rec.) = 1425} \\ \hline

\end{tabular}

\end{center}

\end{table}

All the collected images were labelled with the correct object identity, irrespective of viewing range, angle and receiver height. A fixed size bounding box of $400 \times 400$ cells, which corresponds to $3m^2$,  was cropped from the image with the object in the middle 

\subsection{Neural Network Architecture}
	
We can formalize a neural network as a function with its weights to be learned.

\begin{equation}\label{eq:yfun}
y^l = f(x^l; W^l) 
\end{equation}

where $y$ is the output, $f$ is the neural network function, $W^l$ is a set of weights at layer $l$ and $x^l$ is the input at layer $l$. The neural network needs to be able to learn $W^l$ which will be generalized to any input. The architecture used has several layers; convolutional layers, rectified linear units (ReLU), max pooling, dropout layers and softmax \cite{Goodfellow:2016:DL:3086952}.

\textbf{Convolution Layer:} The main layer developed for deep neural networks when applied to computer vision is the convolutional layer. This learns convolutional masks which are used to extract features based on spatial information. The Eq. \ref{eq:conv} shows the convolution layer computation for each mask.

\begin{equation}\label{eq:conv}
h(i,j,k) = \sum_{u=0}^{M-1} \sum_{v=0}^{N-1} \sum_{w=0}^{D-1} W^l(u,v) . X(i - u, j - v, k - w) + b^l
\end{equation}

In Eq. \ref{eq:conv}, $M$ is the mask width, $N$ is the mask height, $D$ is the mask depth, $W^l$ is the convolution mask learned, $b^l$ is the bias and $X$ is the image.

\textbf{Rectified Linear Unit:} The activation function $f$ is usually a non-linear function that maps the output of current layer. A simple method that is computationally cheap and approximates more complicated non-linear functions, such as, $tanh$ and $sigmoid$, is the Rectified Linear Unit (ReLU).

\begin{gather}\label{eq:fgfun}
f(X) = max(0, X)
\end{gather}

Eq. \ref{eq:fgfun} shows the ReLU function where $X$ is the output from the current layer.

\textbf{Max Pooling:} To reduce the image dimensions, max pooling can be used, simply taking a region and extracting the maximum value. It uses the maximum value since these are the values which have a better activation in the previous layers.

\textbf{Dropout}: The dropout technique was introduced in \cite{krizhevsky2012imagenet, srivastava2014dropout}. This technique simply sets random weights to 0 during training, forcing the neural network to find other paths to train the neural network. This technique avoids overfitting.

\textbf{Softmax}: The softmax layer converts the output from a previous layer into pseudo-probabilities. Thus, for each class, it gives the likelihood of a certain class. Eq. \ref{eq:softmax} shows the softmax layer, where $x_i$ is the output for the current class and $x_j$ is the output for all classes. Hence, it normalises the output vector to 1.

\begin{equation}\label{eq:softmax}
    Softmax(x_i) = \frac{e^{x_i}}{\sum_j e^{x_j}}
\end{equation}

\begin{figure}[!]
\centering
\includegraphics[width=0.35\linewidth]{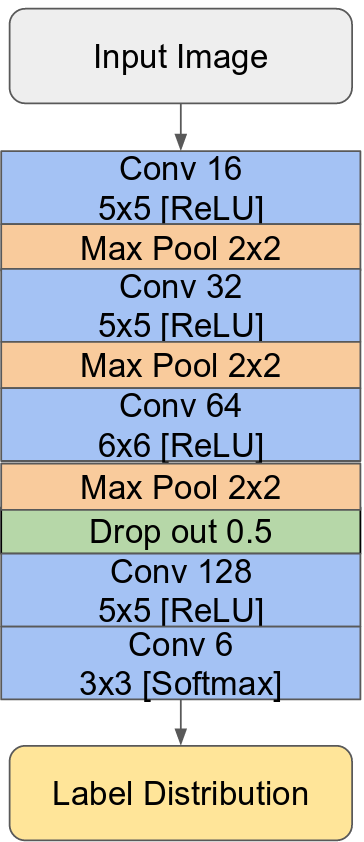}
\caption{A-ConvNet architecture}
\label{fig:aconvnet}
\end{figure}

The neural network used in this work \cite{chen2016target} is A-ConvNet, shown in Figure \ref{fig:aconvnet}. We have  re-coded and implemented this network ourselves from the description given in \cite{chen2016target} using the \texttt{keras} framework \cite{chollet2015keras}.
The only modification we have made is in the last layer with 6 convolutional filters which represents the number of classes for our classification. This architecture is fully convolutional and achieved state-of-the-art results for the MSTAR radar dataset \cite{keydel1996mstar}. The original input A-Convnet is $88 \times 88$, so our input data was re-sized using bilinear interpolation to fit the original model.

There were two principal reasons for using A-ConvNet. First, it achieved excellent results on the radar MSTAR dataset.
Second, we wanted to investigate transfer learning using the same network and sharing the initial weights. Our intuition was that such transfer was more likely to be successful in images of the same modality, i.e. radar, even though their sensing specifications were markedly different.

To train our neural network, we used Stochastic Gradient Descent (SGD). SGD updates the weights of the network depending on the gradient of the function that represents the current layer, as in Eq. \ref{eq:sgd}.

\begin{equation}\label{eq:sgd}
W_{t+1} = W_{t} - \alpha \nabla f(x;W) + \eta \Delta W 
\end{equation}

In Eq. \ref{eq:sgd}, $\eta$ is the momentum, $\alpha$ is the learning rate, $t$ is the current time step, $W$ defines the weights of the network and $\nabla f(x;W)$ is the derivative of the function that represents the network. To compute the derivative for all layers, we need to apply the chain rule, so we can compute the gradient through the whole network. The loss function used to minimise was the categorical cross-entropy (Eq. \ref{eq:loss}). The parameters used in all experiments in all training procedures are given in Table \ref{tab:nn_params}. For all experiments we used 20\% of the training data as validation, and we used the best results from the validation set to evaluate the performance.
In Eq. \ref{eq:loss} $\hat y$ is the predicted vector from softmax output and $y$ is the ground truth.

\begin{equation}\label{eq:loss}
    L(\hat y,y) = - \sum_i y_i log(\hat y_i)
\end{equation}

\begin{table}[!]
\caption{Neural Network Parameters.}\label{tab:nn_params}
\centering

\begin{tabular}{l|l}
\hline
Learning rate ($\alpha$) & 0.001 \\
Momentum ($\eta$)      & 0.9   \\
Epochs        & 100   \\
Batch size    & 100  \\
\hline
\end{tabular}

\end{table}

\subsection{Data Augmentation}

As shown in Table \ref{tab:tab_dataset}, we have limited training data.
Using a restricted dataset, the DCNNs will easily overfit and be biassed towards specific artifacts in the dataset. To help overcome this problem, we generated new samples to create a better generalisation. The simple technique of random cropping takes as input the image data of size $128 \times 128$ and creates a random crop of $88 \times 88$.
This random crop ensures that the target is not always fixed at the same location, so that the location of object should not be a feature. We cropped each sample 8 times and also flipped all the images left to right to increase the size of the dataset and remove positional bias.

\section{Experiments: classification of isolated objects}
\label{sec:experiments}

 As described in Section \ref{sec:data_collected}, we used six objects imaged from ninety viewpoints with three receivers at two different ranges (3.8 m and 6.3 m).
 Four different experiments performed, shown in Table \ref{tab:experiments}. 
In each experiment we compared the results with and without Transfer Learning (TL) from MSTAR. In all training scenarios, data augmentation using random crops and image mirroring from the original data were performed. The metric used to evaluate the results is accuracy, i.e. the number of correct divided by the total number of classifications in the test data. 

\begin{table}[htbp]
\caption{Set of Experiments performed.}\label{tab:experiments}
\centering
\begin{tabular}{lll}
\hline
	& \textbf{Train}	& \textbf{Test}\\
\hline
Experiment 1		& Random (70 \%)			& Random (30 \%)\\
Experiment 2		& 2 Receivers			& 1 Receiver\\
Experiment 3		& One Range		& Other range \\
Experiment 4		& Quadrants 1,3	& Quadrants 2,4\\

\hline
\end{tabular}

\end{table}

\subsection*{Experiment 1: Random selection from the entire data set}

This is the often used, best case scenario, with random selection from all available data to form training and test sets. Intuitively, the assumption is that the dataset contains representative samples of all possible cases. To perform this experiment we randomly selected 70 \% of the data as training and 30 \% as test data. The results are summarised in Table \ref{tab:exp1}



\begin{table}[htbp]
\centering
\caption{Accuracy for experiment 1.}
\label{tab:exp1}
\begin{tabular}{l|c}
\hline
Random Selection from All Data & 99.7\% \\

\hline
\end{tabular}

\end{table}

From Table \ref{tab:exp1} we conclude that the results are very high across the board, so it is possible to recognize objects in the 300 GHz radar images, with the considerable caveats that the object set is limited, they are at short range in an uncluttered environment, and as all samples are used to train, then any test image will have many near neighbours included in the training data with a high statistical probability.

\subsection*{Experiment 2: Receiver/Height influence}

The second experiment was designed to investigate the influence of the receiver antenna characteristics and height (see Figure \ref{fig:exp_config}).
The potential problem is that the DCNNs may effectively overfit the training data to learn partly the antenna pattern from a specific receiver or a specific reflection from a certain height.
All available possibilities were tried, i.e. 

\begin{itemize}
    \item  \textbf{Experiment 2.1 :} Receiver 2 and 3 to train and receiver 1 to test
    \item  \textbf{Experiment 2.2 :} Receiver 1 and 3 to train and receiver 2 to test
    \item \textbf{Experiment 2.3 :} Receiver 1 and 2 to train and receiver 3 to test
\end{itemize}

Table \ref{tab:exp2} shows the results for experiment 2.
In comparison with Experiment 1, the results are poorer, but not by an extent that we can determine as significant on a limited trial. This was expected from examination of the raw radar data, since there is not much difference in the signal signatures from the receivers at different heights. If anything, receiver 3, which was closest to the floor and so received more intense reflections, gave poorer results when used as the test case which implied that the DCNNs did include some measure of receiver or view-dependent characteristics from the learnt data.

\begin{table}[htbp]
\caption{Accuracy for Experiment 2: Receiver Influence}\label{tab:exp2}
\centering
\begin{tabular}{l|c}
             \hline
Receiver 1 Test Experiment & 98.9\% \\ 
Receiver 2 Test Experiment & 98.4\% \\
Receiver 3 Test Experiment & 87.7\% \\
\hline
\end{tabular}

\end{table}
    
\subsection*{Experiment 3: Range influence}

Clearly, the range of the object influences the return signature to the radar as the received power will be less due to attenuation, and less cells are occupied by the target in the radar image due to degrading resolution over azimuth. Therefore, if the training data set is selected only at range 3.8m. for example, to what extent are the features learnt representative of the expected data at 6.8m (and vice versa)? Table \ref{tab:exp3} summarises the results achieved when we used one range to train the network, and the other range to test performance.

\begin{itemize}
    \item \textbf{Experiment 3.1 :} Train with object on 3.8 m. Test with object on 6.3 m.
    \item \textbf{Experiment 3.2 :} Train with object on 6.3 m. Test with object on 3.8 m.
\end{itemize}

    \begin{table}[htbp]
    \caption{Accuracy for Experiment 3: Range Influence}\label{tab:exp3}
    \begin{center}
    \centering
    \begin{tabular}{l|c}
    \hline
Object at 6.3 m Test Experiment & 82.5\% \\
Object at 3.8 m Test Experiment & 91.1\%  \\

    \hline
    \end{tabular}
    \end{center}
    
    \end{table}
    
The key observation from Table \ref{tab:exp3} is that if we train the DCNNs at one specific range which has a given cell structure and received power distribution, and then test at a different range, the DCNNs is not as accurate as the base case as this drops from over 99\% to 82\% and 91\% respectively.

\subsection*{Experiment 4: Orientation influence}

\begin{figure}[h]
    \centering
    \includegraphics[width=0.3\linewidth]{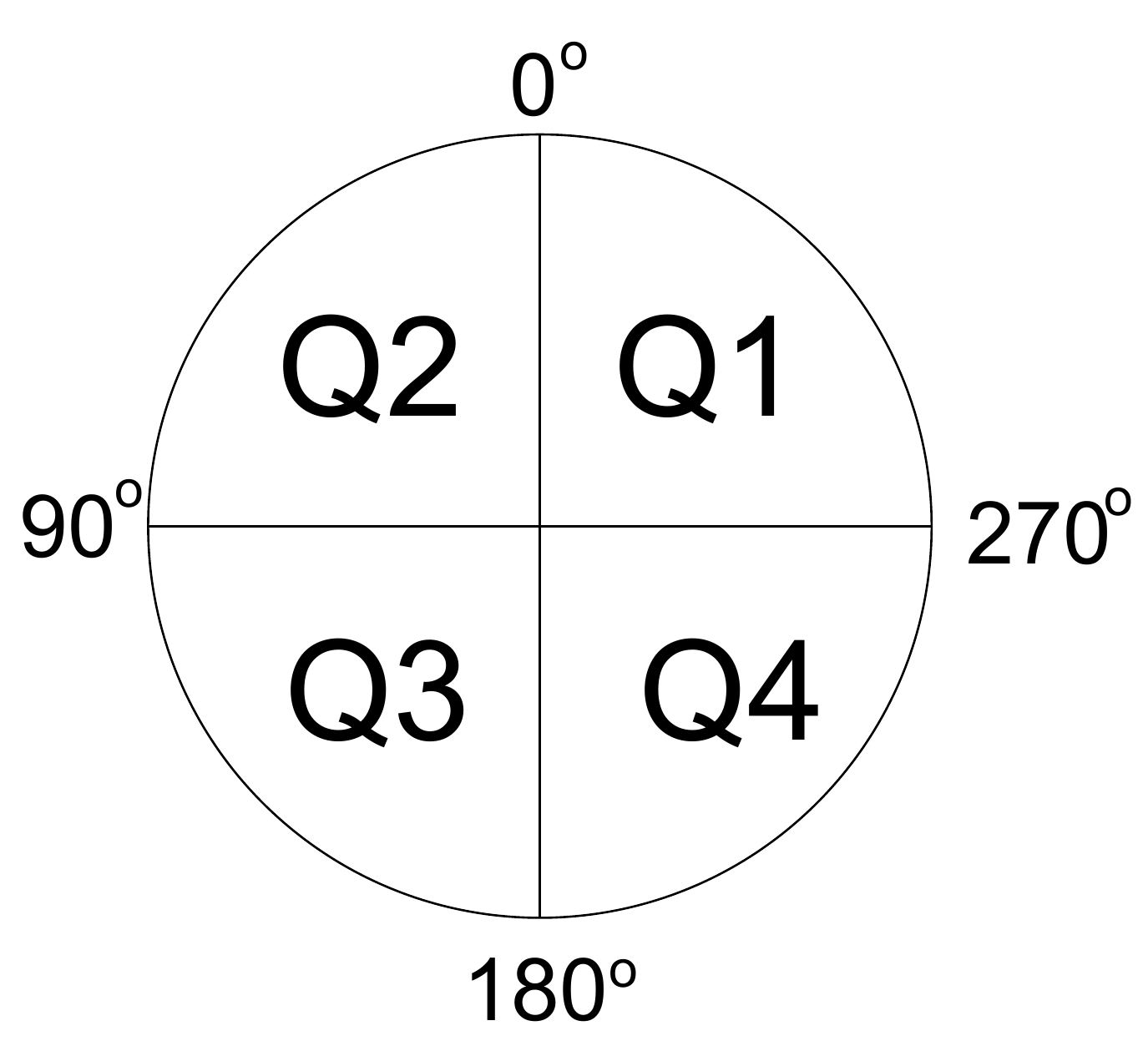}
    \caption{Quadrants}
    \label{fig:quadrant}
\end{figure}

The final experiment was designed to examine whether the neural network was robust to change of viewing orientation. Here, we used as training sets the objects in quadrants 1 and 3, and as test sets the objects in  quadrants 2 and 4. Quadrant 1 means orientation from $0^o$ to $89^o$, quadrant 2 means orientation from $90^o$ to $179^o$, quadrant 3 means orientation from $180^o$ to $269^o$ and quadrant 4 means orientation from $270^o$ to $359^o$, as seen in the Figure \ref{fig:quadrant}

    \begin{table}[htbp]
    \caption{Accuracy for experiment 4: Orientation influence}\label{tab:exp4}
    \centering
    \begin{tabular}{l|c}
    \hline
Q2,Q4 Test Experiment &92.5\% \\
\hline

    \end{tabular}
    
    \end{table}

The DCNNs does not perform as well compared to Experiments 1 and 2, dropping to 92.5\%.
However, since we flipped the images left to right as a data augmentation strategy, the network was capable of learning the orientation features, as the objects exhibit near mirror symmetry, and in one case, the cone, is identical from all angles. 
Therefore, we have to be hesitant in drawing conclusions about any viewpoint invariance within the network as the experiments are limited and all objects have an axis or axes of symmetry (as do many objects in practice).

Together with Experiments 2 and 3, this experiment shows that it is necessary to take into account the differences in the acquisition process using different receivers at different ranges and orientation in training the network.
While, this is to some extent obvious and equally true for natural images, we would observe that the artefacts introduced by different radar receivers are much less standardised that those introduced by standard video cameras, so the results obtained in future may be far less easy to generalise. Although Experiment 2 only showed limited variation in such a careful context, we would speculate that the effects of multipath and clutter would be far more damaging than in the natural image case, as highlighted in \cite{bartsch2012pedestrian}.

\subsection{Experiments using Transfer Learning}

\subsection{Transfer Learning}
	
As summarised in the previous Section, we have a small dataset and there is the potential to learn characteristics of the restricted dataset rather than of the objects themselves.
Therefore, we have investigated the use of transfer learning to help capture more robust features using a pre-existing dataset, i.e. to use prior knowledge from one domain and transfer it to another \cite{yosinski2014transferable}.
To apply transfer learning, we first trained the DCNNs on the MSTAR (source) data, then the weights from the network were used as initial weights for a new DCNNs trained on our own 300 GHz (target) data.

\begin{figure}[t!]
\centering
\includegraphics[width=\linewidth]{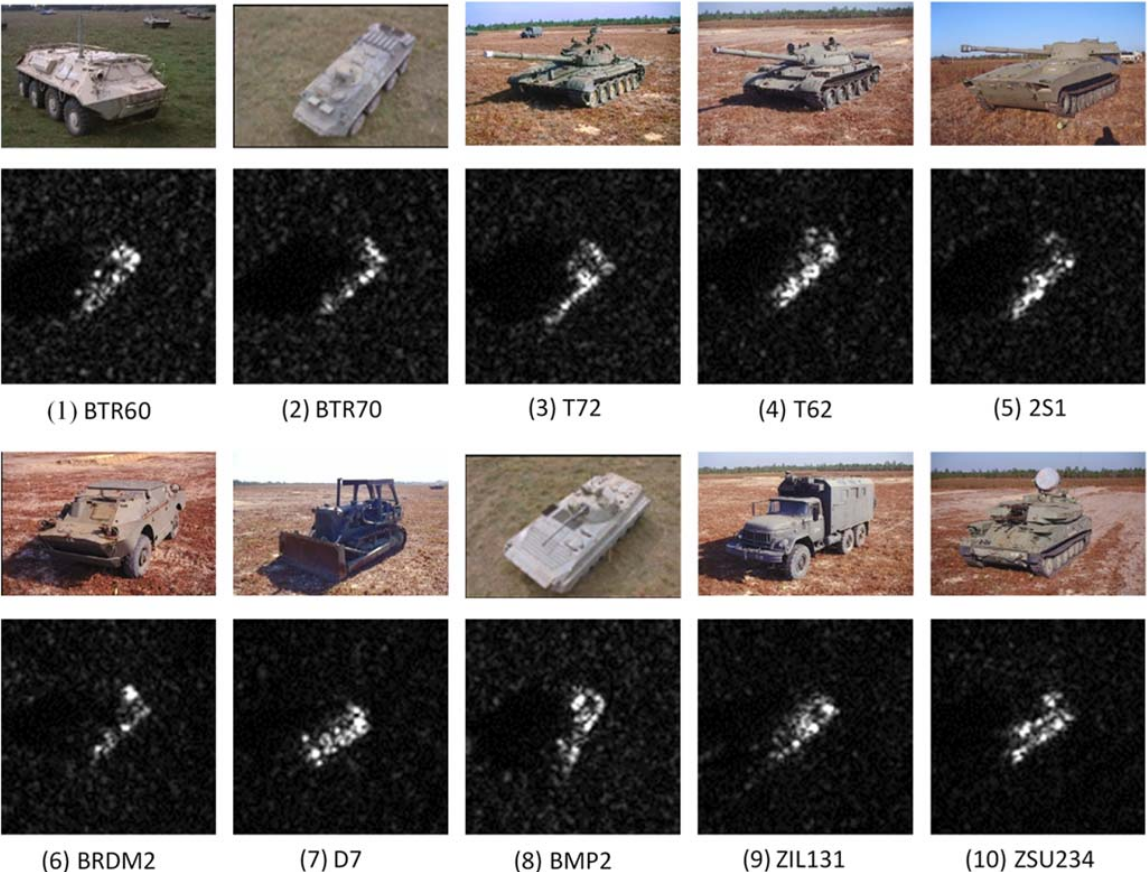}
\caption{MSTAR Dataset}
\label{fig:mstar_dataset}
\end{figure}

The MSTAR data is different in viewing angle and range compared to our own data as shown in Figure \ref{fig:mstar_dataset}.
It was developed to recognise military targets using SAR images. The data contains 10 different military targets and around 300 images per target with similar elevation viewing angles of \ang{15} and \ang{17}.
In total MSTAR has around 6000 images and is used widely by the radar community in order to verify classification algorithms.

The DCNNs function in the source domain is defined by Eq. \ref{eq:dnn}.
	
\begin{equation}\label{eq:dnn}
y_s = f(W_s,x_s)
\end{equation}
	
\noindent 
where $W_s$ are the weights of a network, $x_s$ and $y_s$ are the input and and output from the source domain. To learn the representation, an optimizer must be used, again stochastic gradient descent (SGD), expressed in Eq. \ref{eq:optimizer}.
	
\begin{equation}\label{eq:optimizer}
W_{s_{i+1}} = SGD(W_{s_i}, x_s, y_s)
\end{equation}
	
\noindent 
where $SGD$ is a function which updates the weights of the neural network, as expressed in Eq. \ref{eq:sgd}
Hence, using the trained weights from our source domain as the initial weights, this is expressed as Eq. \ref{eq:tf}.
It is intended that the initial weights give a better initial robust representation which can be adapted to the smaller dataset. 
	
	\begin{equation}\label{eq:tf}
	W_{t_1} = SGD(W_s,x_t, y_t), \textnormal{when i = 0}
	\end{equation}

We repeated experiments 1,2,3 and 4 using transfer learning. The results are summarised in Table \ref{tab:exp_tl}.

\begin{figure*}[t]
    \centering
    \begin{subfigure}[b]{0.3\textwidth}
        \includegraphics[width=\textwidth]{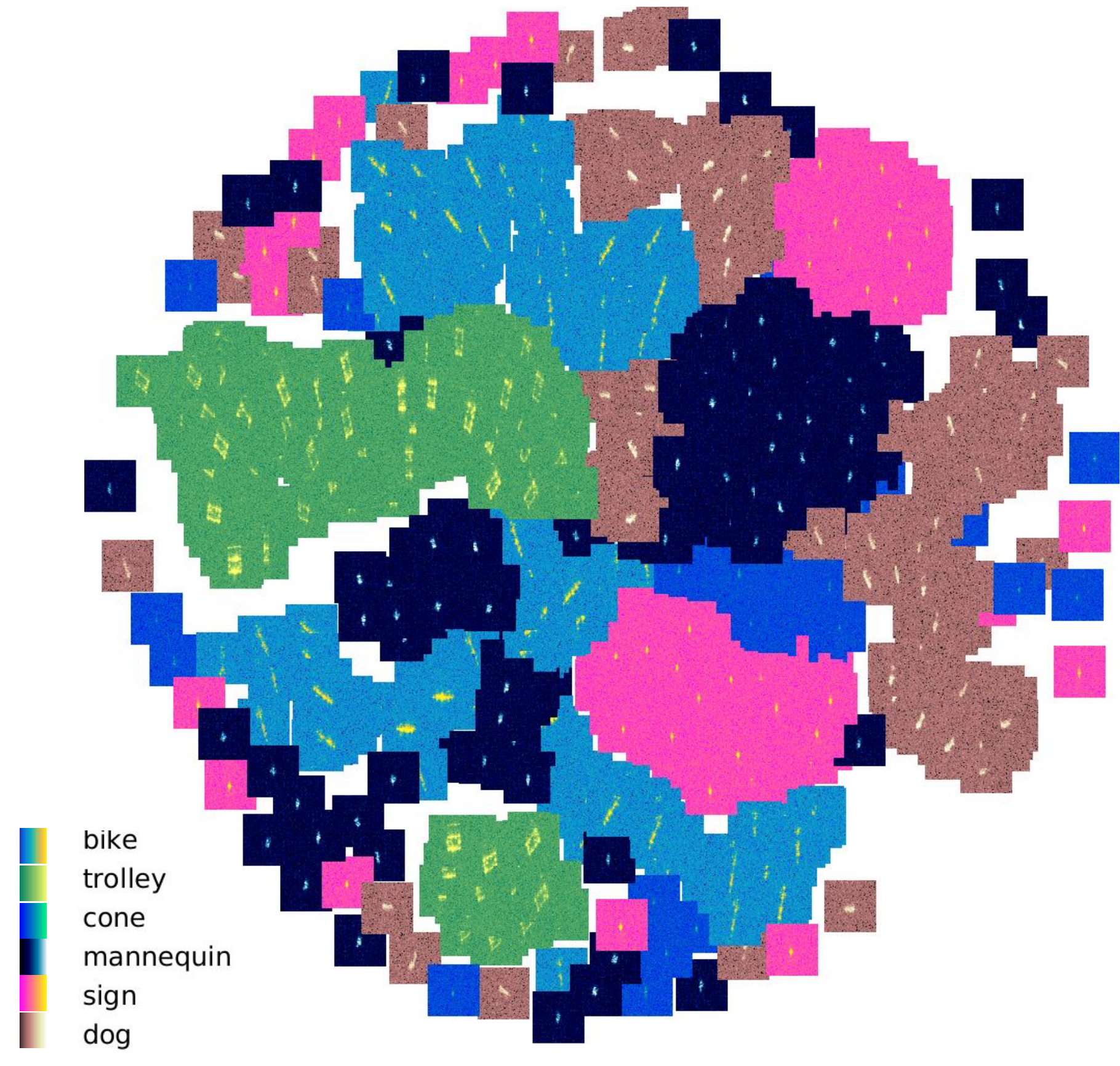}
        \caption{t-SNE using raw features}
        \label{fig:tsne_raw}
    \end{subfigure}%
    \begin{subfigure}[b]{0.3\textwidth}
        \includegraphics[width=\textwidth]{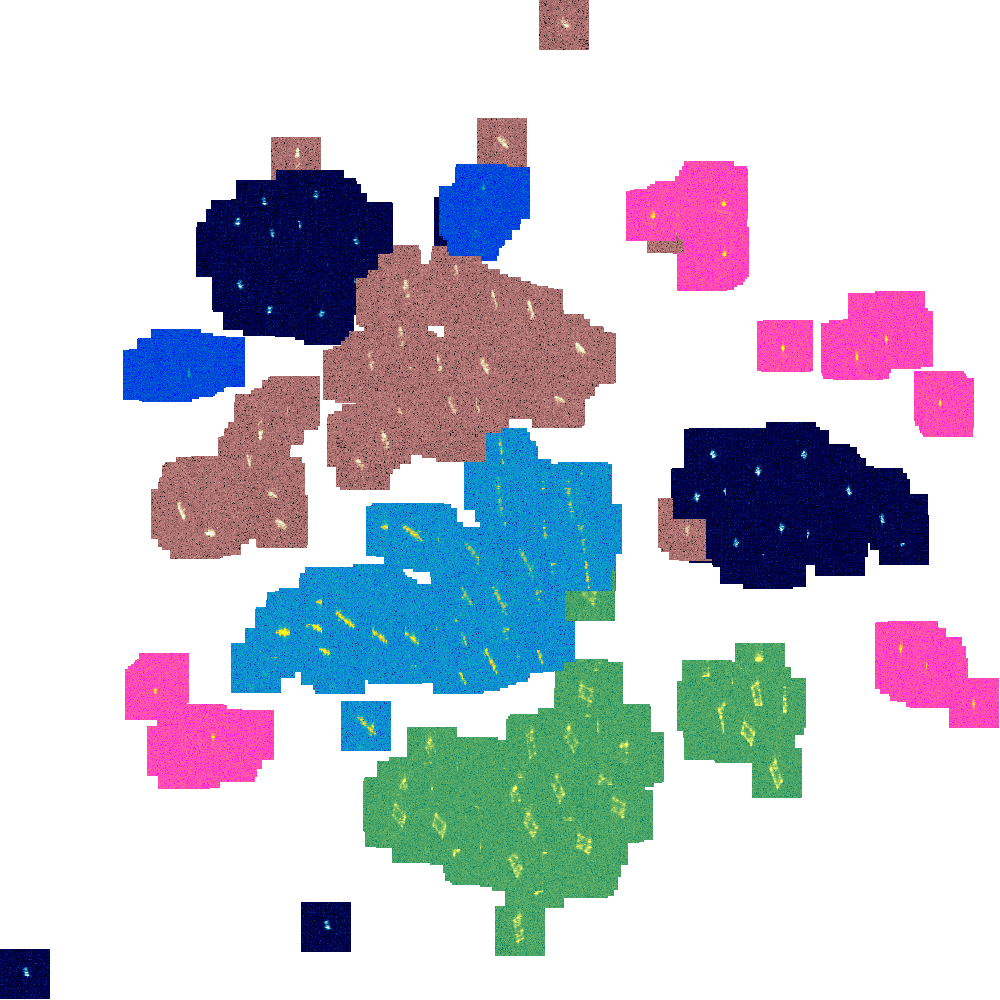}
        \caption{t-SNE without Transfer learning}
        \label{fig:tsne_da}
    \end{subfigure}%
    \begin{subfigure}[b]{0.3\textwidth}
        \includegraphics[width=\textwidth]{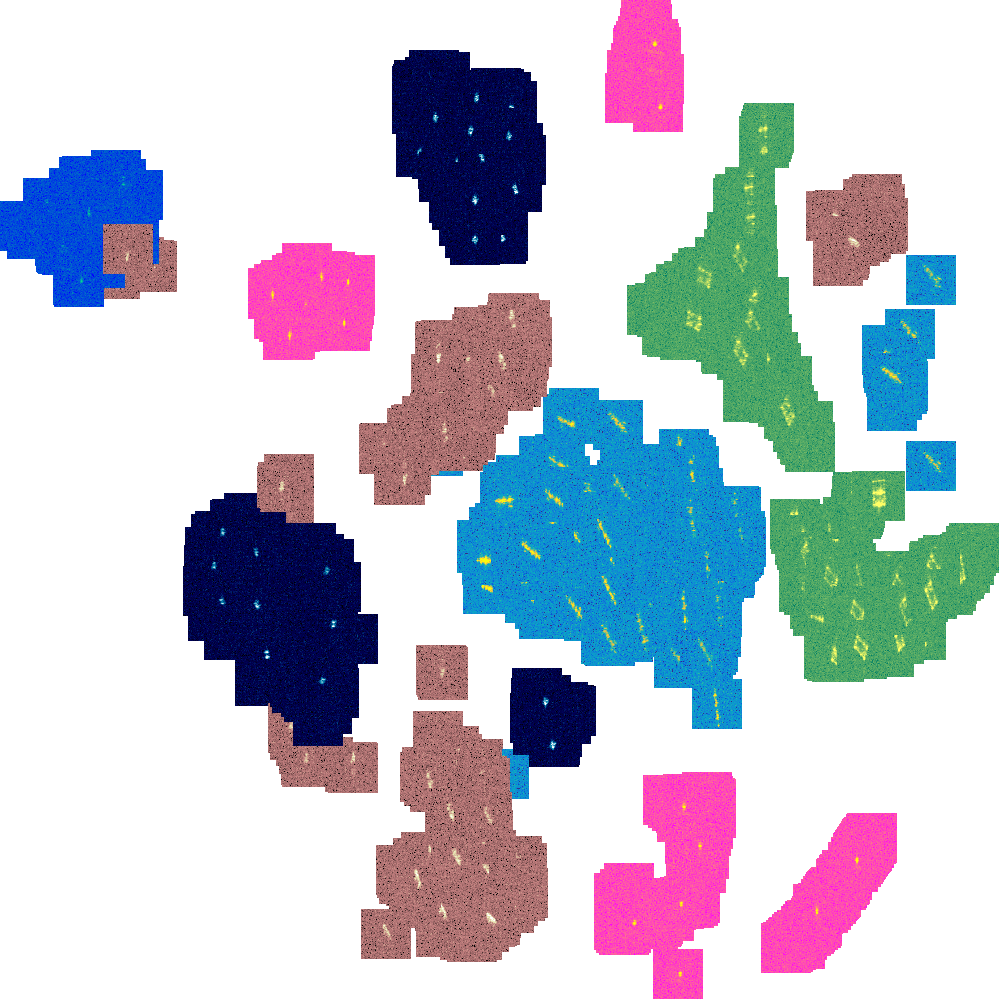}
        \caption{t-SNE with Transfer learning}
        \label{fig:tsne_da_tl}
    \end{subfigure}
    \caption{t-SNE plots from the orientation experiment}\label{fig:tsne_plots}
\end{figure*}
    
    \begin{table}[htbp]
        \caption{Accuracy after applying transfer learning}\label{tab:exp_tl}
    \centering
    \begin{tabular}{l|cc}
    \hline
    
    & without TL & with TL \\
    \hline
Random Split Exp. & \textbf{99.7\%} & 99.1\% \\
Rec. 1 Test Exp. & \textbf{98.9\%} & 95.8\% \\
Rec. 2 Test Exp. & 98.4\% & \textbf{98.8\%} \\
Rec. 3 Test Exp. & 87.7\% & \textbf{94.1\%} \\
6.3 m Test Exp. & 82.5\% & \textbf{85.2\%} \\
3.8 m Test Exp. & 91.1\% & \textbf{93.5\%} \\
Q2,Q4 Test Exp. & 92.5\% &\textbf{98.5\%}  \\
    \hline
    \end{tabular}

    \end{table}
    
As can be seen, transfer learning gives higher values for accuracy in the majority but not all cases. The MSTAR dataset is a much bigger dataset, and although it exhibits some characteristics in common with our own data, it uses a synthetic aperture technique, and there is no significant variation in elevation angle during data collection.  However, there are 2 distinguishable strong features, the shape and reflected power. 
As these have much in common with our own data, it is possible that the network is able to better generalise to cope with new situations as shown in the the Receiver 3 and different range experiments for example.
To draw any firmer conclusion requires much more extensive evaluation.

However, to gain further insight, We also show the confusion matrix from the orientation experiments without and with transfer learning in Tables. \ref{tab:conf_mat} and \ref{tab:conf_mat_tl}. The main confusion is between the dog and mannequin, since both have similar clothed material; and cone and sign, since they have similar shape. Nevertheless, in these experiments, we can conclude that the neural network approach is robust in maintaining accuracy with respect to sensor hardware, height, range and orientation.

    
\begin{table}[]
\caption{Orientation Experiment without Transfer Learning}\label{tab:conf_mat}
\centering
\begin{tabular}{ll|cccccc}
 \multicolumn{2}{c|}{\multirow{2}{*}{Acc: 0.925}} &\multicolumn{6}{c}{Predicted Label} \\
 &           & Bike                 & Trolley              & Cone                 & Mannequin            & Sign                 & Dog                  \\ \hline
 \parbox[t]{2mm}{\multirow{6}{*}{\rotatebox[origin=c]{90}{True Label}}}& Bike      & 1.00                 &   0.00                   &     0.00                 &         0.00             &        0.00              &         0.00             \\
 & Trolley   &    0.03                  & 0.97                 &    0.00                  &       0.00               &   0.00                   &     0.00                 \\
 & Cone      &     0.00                 &       0.00               & 1.00                 &      0.00                &    0.00                  &    0.00                  \\
 & Mannequin &      0.00                &   0.00                   &     0.00                 & 0.86                 &   0.00                   & 0.14                 \\
 & Sign      &       0.00               &    0.00                  &    0.00                  &       0.00               & 1.00                 &   0.00                   \\
 & Dog       & 0.03                 &              0.00        & 0.02                 & 0.10                 &    0.00                  & 0.86                
\end{tabular}
\end{table}

\begin{table}[]
\caption{Orientation Experiment with Transfer Learning}\label{tab:conf_mat_tl}
\centering
\begin{tabular}{ll|cccccc}
 \multicolumn{2}{c|}{\multirow{2}{*}{Acc: 0.985}} &\multicolumn{6}{c}{Predicted Label} \\
 &           & Bike                 & Trolley              & Cone                 & Mannequin            & Sign                 & Dog                  \\ \hline
 \parbox[t]{2mm}{\multirow{6}{*}{\rotatebox[origin=c]{90}{True Label}}}& Bike      & 1.00                 &   0.00                   &     0.00                 &         0.00             &        0.00              &         0.00             \\
 & Trolley   &    0.00                  & 1.00                 &    0.00                  &       0.00               &   0.00                   &     0.00                 \\
 & Cone      &     0.00                 &       0.00               & 1.00                 &      0.00                &    0.00                  &    0.00                  \\
 & Mannequin &      0.00                &   0.00                   &     0.00                 & 0.96                 &   0.00                   & 0.04                 \\
 & Sign      &       0.00               &    0.00                  &    0.00                  &       0.00               & 1.00                 &   0.00                   \\
 & Dog       & 0.00                 &              0.00        & 0.00                 & 0.03                 &    0.00                  & 0.97                
\end{tabular}
\end{table}

\subsection{Visualisation of feature clusters}

To better understand what is being learned by our network, the t-SNE technique \cite{maaten2008visualizing} was used to visualise the feature clusters. t-SNE employs nonlinear dimensionality reduction to build a probability distribution by comparing the similarity of all pairs of data, then transformed to a lower dimension. Then it uses KL-divergence to minimise with respect to the locations in the cluster space.

Figure \ref{fig:tsne_plots} shows the result from t-SNE clustering of samples using raw image features, in this case the orientation experiment. Figures \ref{fig:tsne_da} and \ref{fig:tsne_da_tl} show the t-SNE clusters from the features extracted from the penultimate layer of the trained neural network with and without transfer learning, using different colormaps for each object for better visualisation. We can see that the trained neural network was able to cluster similar classes and similar features. It is hard to give actual interpretability of neural networks, the t-SNE framework can give some insights of the type of features that have been learned. The transfer learning cluster shows slight improvement by creating bigger clusters of objects of the same class.

\section{Experiments: Detection and classification within a multiple object scenario}
\label{sec:detection}

\begin{figure*}[h]
\centering
\includegraphics[width=\linewidth]{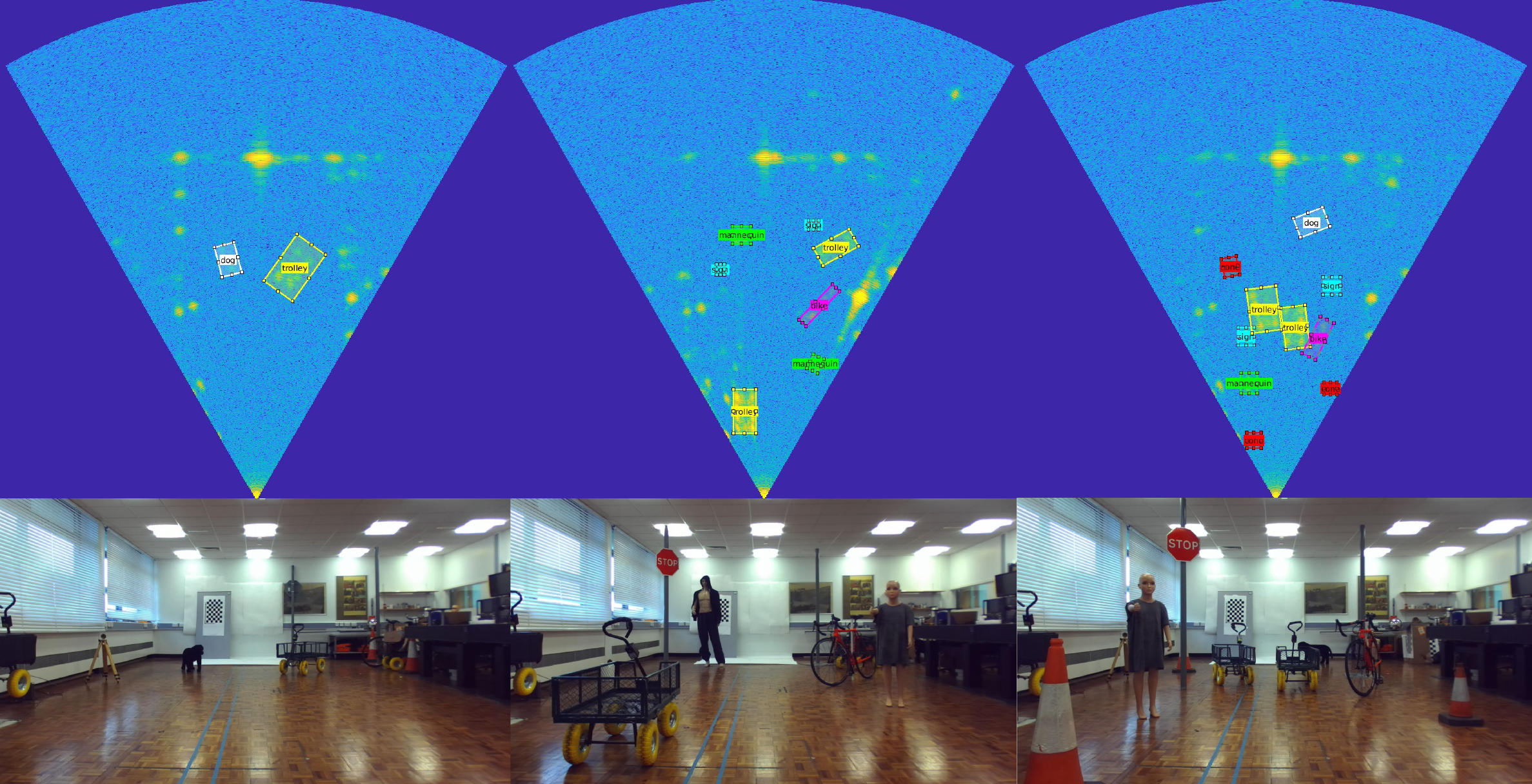}
\caption{Multiple Object Dataset. Above: 300 GHz radar image. Below: Reference RGB image.}
\label{fig:multi_object}
\end{figure*}

\begin{figure*}[h]
\centering
\includegraphics[width=0.7\linewidth]{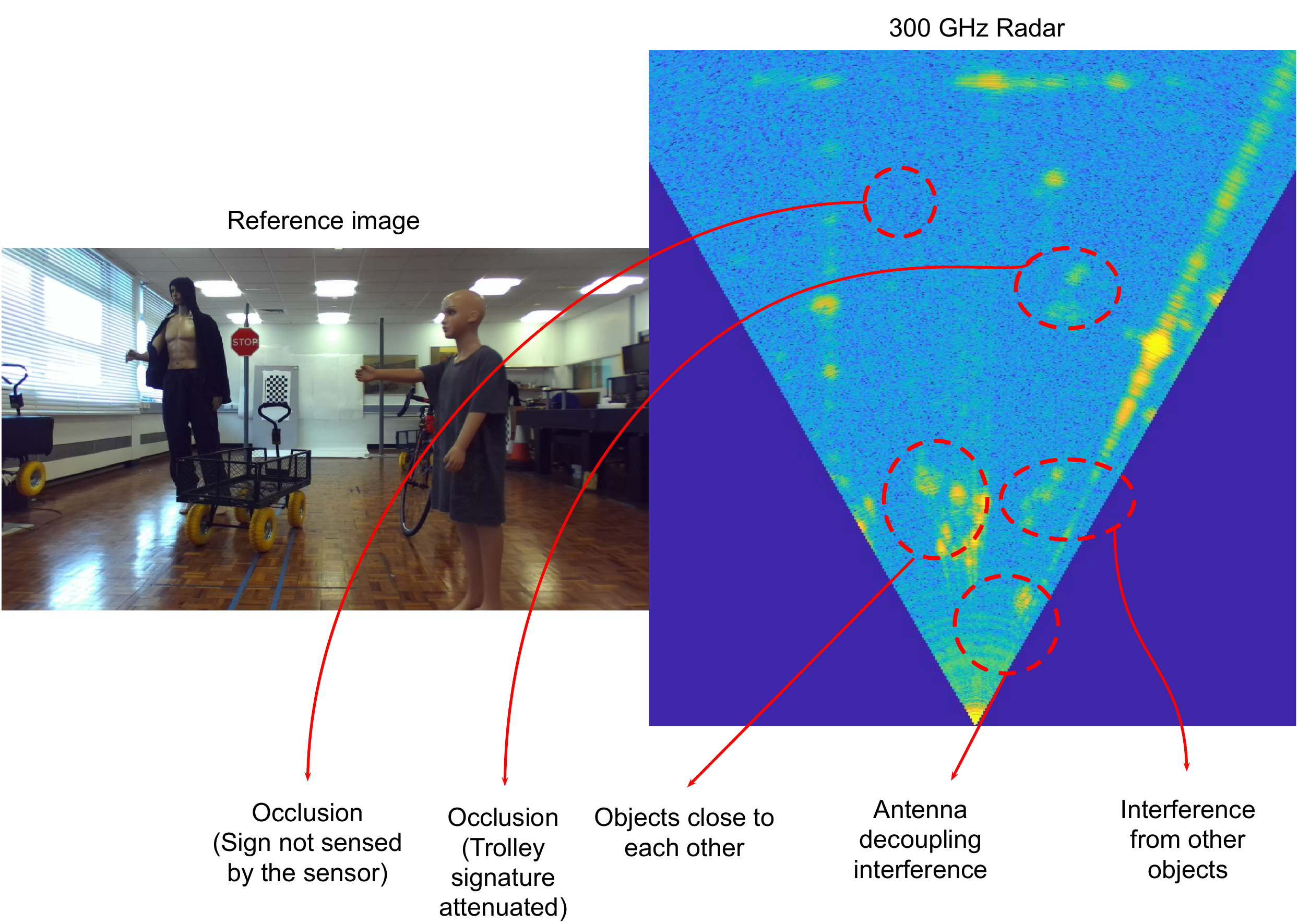}
\caption{Possible unwanted effects in the multiple object dataset}
\label{fig:multi_object_problems}
\end{figure*}

\begin{figure*}[t]
\centering
\begin{subfigure}{.3\textwidth}
  \centering
  \includegraphics[width=\linewidth]{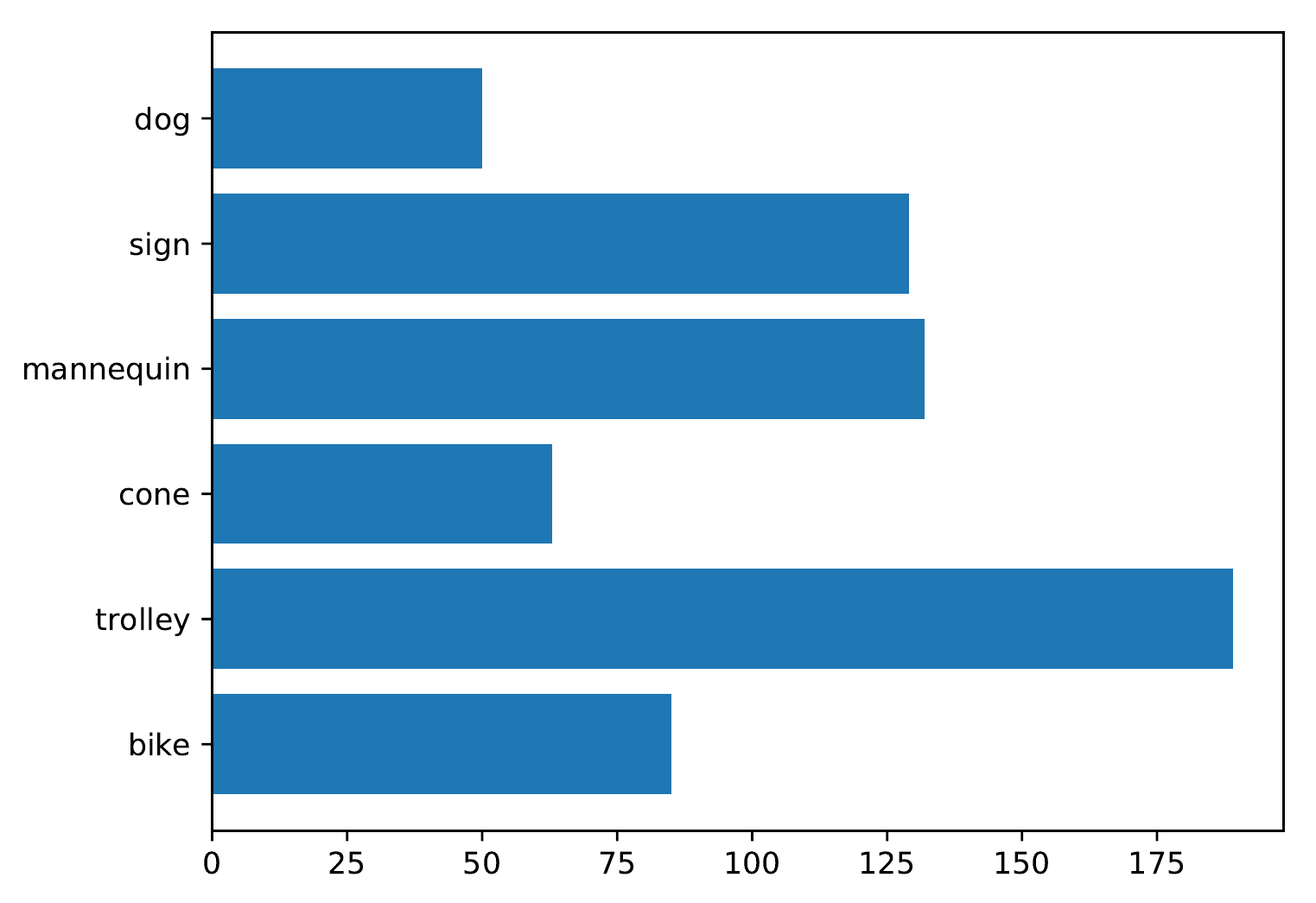}
  \caption{Number of objects per class}
  \label{fig:nb_obj}
\end{subfigure}%
\begin{subfigure}{.3\textwidth}
  \centering
  \includegraphics[width=\linewidth]{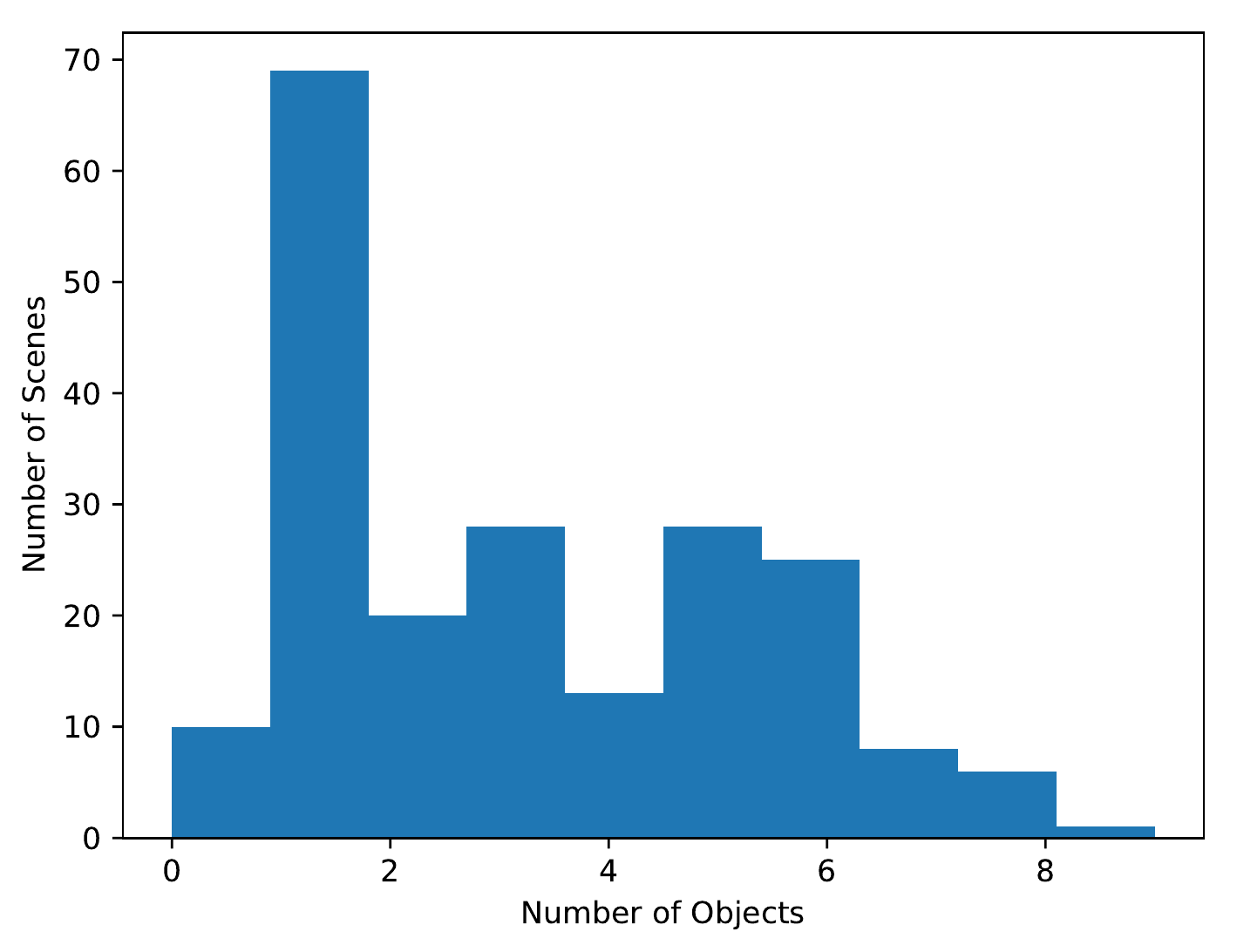}
  \caption{Histogram of number of objects per scene}
  \label{fig:hist_obj}
\end{subfigure}
\begin{subfigure}{.3\textwidth}
  \centering
  \includegraphics[width=\linewidth]{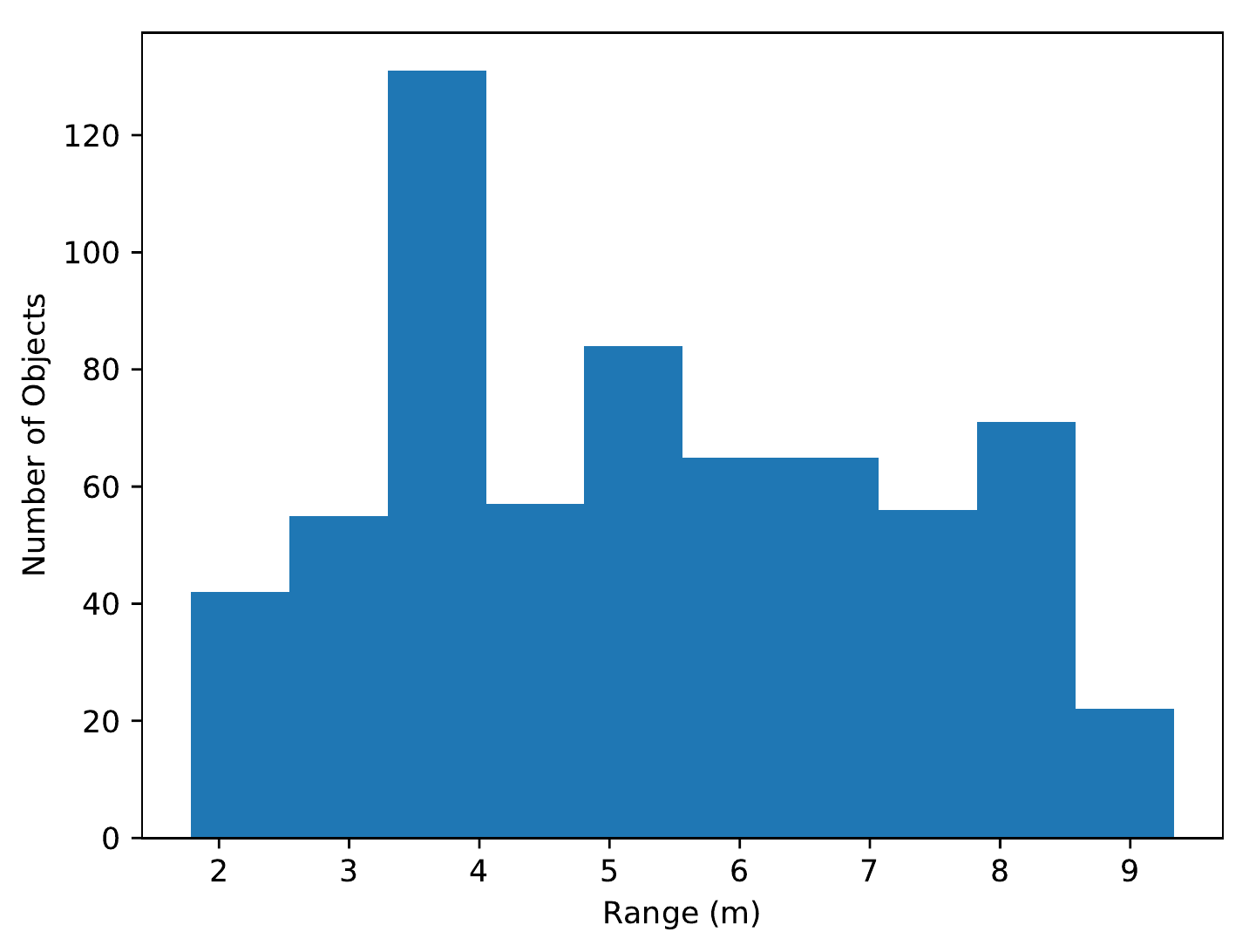}
  \caption{Histogram distribution of ranges}
  \label{fig:hist_range}
\end{subfigure}
\caption{Multi-object dataset statistics}
\label{fig:multiobj_stats}
\end{figure*}

The previous dataset contains one windowed object in each image. In an automotive or more general radar scenario we must both detect and classify road actors in a scene with many pre-learnt and unknown objects which is much more challenging. Hence, in the next set of experiments we include multiple objects, and this has several additional phenomena including occlusion, multi-path and interference between objects, as well as objects which are not included as a learnt object of interest.  We use the same object dataset (bike, trolley, cone, mannequin, sign, dog) in different parts of the room with arbitrary rotations and ranges, and the network is trained by viewing the objects in isolation, as before. We also include some within-object variation, using for example different mannequins, trolleys ad bikes. The unknown, laboratory walls are also very evident in the radar images. This new dataset contains 198 scenes, 648 objects, an average of 3.27 movable objects per scene.
Fig. \ref{fig:multi_object} shows examples of 3 scenes in the multiple object dataset.  Fig. \ref{fig:multiobj_stats} shows statistical data explaining the number of instances of each learnt object, the number of objects in each scene, and the distribution of ranges of the objects.  Fig. \ref{fig:multi_object_problems} illustrates possible problems that can occur in the multiple objects dataset.

        
    

\subsection{Methodology}\label{sec:det_method}

In classical radar terminology, detection is described as "determining whether the receiver output at a given time represents the echo from a reflecting object or only noise" \cite{richards2005fundamentals}.
Conversely, in computer vision, using visible camera imagery to which the vast majority of CNN methods have been applied, detection is the precise location of an object in an image (assuming it is present) containing many other objects, as for example in the pedestrian detection survey of Dollar et al. \cite{dollar2012}. Although the image may be noisy, this is generally not the major cause of false alarms.

The extensive literature on object detection and classification using cameras, e.g. \cite{ren2017faster, liu2016ssd, lin2017focal, redmon2016yolo}, can be grouped into \textit{one-stage} and \textit{two-stage} approaches. In the \textit{one-stage} approach localisation and classification is done within a single step, as with the YOLO \cite{redmon2016yolo}, RetinaNet \cite{lin2017focal} and SSD \cite{liu2016ssd} methods. Using a \textit{Two-stage} approach first where is a need to localise then classify each proposed bounding box, then a classification is performed in that box. R-CNN \cite{girshick2014rich}, Fast R-CNN \cite{girshick2015fast} and Faster R-CNN \cite{ren2017faster} are examples of the \textit{two-stage} approach.

\begin{figure*}[t]
    \centering
    \includegraphics[width=\linewidth]{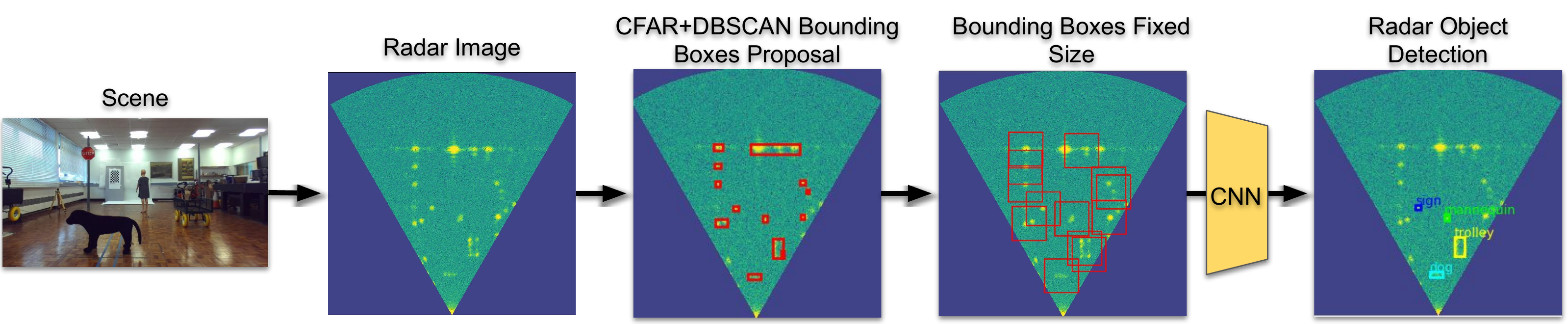}
    \caption{Methodology developed for the detection task.}
    \label{fig:detection_method}
\end{figure*}

For this work we developed a \textit{two-stage} technique. We first generate bounding boxes based on the physical properties of the radar signal, then the image within each bounding box is classified, similar to the R-CNN \cite{girshick2014rich}.
Fig. \ref{fig:detection_method} shows the pipeline of the detection methodology developed.
For radar echo detection, we use simply \textit{Constant False Alarm Rate} (CFAR) \cite{richards2005fundamentals} detection.
There are many variations including \textit{Cell Averaging Constant False Alarm Rate} (CA-CFAR) and \textit{Order Statistics Constant False Alarm Rate} (OS-CFAR). In this work we used the CA-CFAR algorithm to detect potential radar targets. In order to compute the false alarm rate, we measured the background noise level, and the power level from the objects, setting a CFAR level of $0.22$. 
After detecting potential cells, we form clusters using the common \textit{Density-based spatial clustering of applications with noise} (DBSCAN) algorithm \cite{Ester:1996:DAD:3001460.3001507} which forms clusters from proximal points and removes outliers.
For each cluster created we use the maximum and minimum points to create a bounding box of the detected area. The parameters for DBSCAN used were selected empirically; $\epsilon = 0.3 m$ which is the maximum distance of separation between 2 detected points, and $S = 40$, were $S$ is the minimum number of points to form a cluster.

To compute the proposed bounding boxes with DBSCAN, we use the center of the clusters to generate fixed size bounding boxes of known dimensions, since, in contrast to the application of CNNs to camera data, the radar images are metric and of know size. Hence, the boxes are of size $275 \times 275$, the same size as the data used to train the neural network for the classification task. The image is resized to $88 \times 88$ and each box is classified. 

To consider the background we randomly cropped 4 boxes which do not intersect with the ground truth bounding boxes containing objects in each  scene image from the multiple object dataset and incorporated these in our training set. However, as there are effectively two types of background, that which contains other unknown objects such as the wall, and the floor areas which have low reflected power, we ensured that the random cropping contained a significant number of unknown object boxes. This is not ideal, but we are limited to collect data in a relatively small laboratory area due to the restricted range of the radar sensor and cannot fully model all possible cluttering scenarios.


\subsection{Results for Multiple Objects}\label{sec:det_results}

In order to evaluate performance, we have considered 3 different scenarios. 
In particular, we wish to ascertain how performance is affected by failures in classification  assuming a perfect CFAR+DBSCAN pipeline, and to what extent failures in the box detection process lead to mis-classification. Further, we make a distinction between confusing objects (mainly the lab wall) and due to system noise from the floor area.

\begin{itemize}
    \item \textbf{Perfect Detector} : In this scenario we do not use the CFAR + DBSCAN pipeline, we use the ground truth as the detected bounding boxes. Each bounding box is fed to the trained neural network.
    
    \item \textbf{Easy} : In this scenario we manually crop the walls and focus on the potential area containing objects of interest. This includes the CFAR + DBSCAN in a easy scenario, in which removal of static objects is analogous to background subtraction..
    
    \item \textbf{Hard} : In this scenario we assume the whole scene has potential targets. Hence, the wall should result in positive detections and is a challenge to the CNN classification.
\end{itemize}

We also decided to label our scene data depending on the density of objects, since highly cluttered scene should increase the likelihood of unwanted radar sensing effects, such a multi-path, occlusion, and multiple objects in the same bounding box.

\begin{itemize}
    \item \textbf{\#Objects $<$ 4} : At low density of objects, it is likely that the scene will suffer less from these effects.
    \item \textbf{4 $\leq$ \#Objects $<$ 7} : At mid density, we will encounter some of the unwanted effects.
    \item \textbf{\#Objects $\geq$ 7} : At high density, many of these effects occur.
\end{itemize}

We also have decided to evaluate performance at different ranges.

\begin{itemize}
    \item \textbf{Short Range (Objects $<$ 3.5 m)}: This scenario is not necessarily the easiest since coupling between the the transmitter and receiver happens at this range \cite{8468328}.
    \item \textbf{Mid Range (3.5 m $<$ Objects  7 m)}: This is the ideal scenario, as the objects were learnt within these ranges, and the antenna coupling interference is reduced.
    \item \textbf{Long Range (Objects $>$ 7 m)}: This is the most challenging scenario. At more than 7 meters, most of the objects have low power of return, close to the systemic background.
\end{itemize}

The metric we use for evaluation is average-precision (AP) which is a commonly used standard in the computer vision literature for object detection, classification and localisation \cite{everingham2015pascal} in which the Intersection over Union (IoU) measures the overlap between 2 bounding boxes.
If the overlap is greater than 0.5 and the classification is correct, then this is a true positive. To compute AP we need to compute precision (Eq. \ref{eq:prec}) and recall (Eq. \ref{eq:recall}), where TP is  true positive, FP is a false positive and FN is a false negative.
To compute AP we compute the area under the curve from the precision-recall plot varying the confidence level of the prediction of each bounding box. The AP is computed as shown in the Eq. \ref{eq:ap} where $p$ is precision and $r$ is recall.

\begin{equation}
    Precision = \frac{TP}{TP + FP}
\label{eq:prec}
\end{equation}

\begin{equation}
    Recall = \frac{TP}{TP + FN}
\label{eq:recall}
\end{equation}

\begin{equation}
    AP = \int_{0}^{1} p(r)dr
\label{eq:ap}
\end{equation}

For these experiments we retrained the neural network from the single object dataset using the orientation experiments. For the \textit{Easy} and \textit{Perfect Detector} cases, we do not include the background data. For the \textit{Hard} case we also added 4 background images per scene inside our training set.
Extensive results for all these scenarios are shown in Tables \ref{tab:pd} , \ref{tab:dbscan_easy}, \ref{tab:dbscan_hard}.

\begin{table*}[h!]
\caption{Perfect Detector}\label{tab:pd}
\resizebox{\textwidth}{!}{%
\begin{tabular}{l|c|c:ccc|c:ccc|c:ccc|c|c|c}
\multicolumn{1}{c|}{\multirow{2}{*}{AP}}& \multirow{2}{*}{Overall} & \multicolumn{4}{c|}{$\# Objects < 4$} & \multicolumn{4}{c|}{$4 \leq \# Objects < 7$} & \multicolumn{4}{c|}{$\# Objects \geq 7$} & \multirow{2}{*}{Short} & \multirow{2}{*}{Mid} & \multirow{2}{*}{Long} \\
& & Overall & Short & Mid & Long & Overall & Short & Mid & Long & Overall & Short & Mid & Long & & & \\ \hline

bike & 64.88 & 79.17 & 50.00 & 83.33 & 75.00 & 48.42 & 25.00 & 56.77 & 33.33 & 76.26 & N/A & 67.05 & N/A & 35.00 & 66.19 & 57.14 \\ 
cone & 46.87 & 50.00 & 50.00 & 50.00 & 50.00 & 58.29 & 55.56 & 83.33 & N/A & 42.49 & 68.75 & 26.67 & N/A & 62.07 & 43.30 & 3.57 \\ 
dog & 51.34 & 77.62 & 77.78 & 87.72 & 47.62 & 49.13 & 77.78 & 55.45 & 33.33 & 26.40 & 60.0 & N/A & 12.50 & 70.95 & 65.02 & 20.19 \\ 
mannequin & 37.73 & 70.53 & 53.33 & 85.71 & 33.33 & 25.57 & 36.36 & 30.00 & 8.00 & 37.78 & 14.29 & 50.00 & 22.35 & 33.08 & 48.72 & 13.61 \\ 
sign & 85.64 & 81.86 & 0.00 & 89.47 & 66.67 & 86.60 & N/A & 90.10 & 81.08 & 86.44 & N/A & 88.89 & 85.46 & 0.00 & 89.65 & 81.94 \\ 
trolley & 81.68 & 87.75 & 79.17 & 97.06 & 82.35 & 85.35 & 100.00 & 87.61 & 70.13 & 75.45 & 92.67 & 83.65 & 10.00 & 93.53 & 88.76 & 60.41 \\ 
\hline 
mAP & 61.36 & 74.49 & 51.71 & 82.22 & 59.16 & 58.89 & 58.94 & 67.21 & 37.65 & 57.47 & 58.93 & 63.25 & 26.06 & 49.1 & 66.94 & 39.48 \\ 

\end{tabular}}
\end{table*}

\begin{table*}[h!]
\caption{CFAR+DBSCAN Detector Easy}\label{tab:dbscan_easy}
\resizebox{\textwidth}{!}{%
\begin{tabular}{l|c|c:ccc|c:ccc|c:ccc|c|c|c}
\multicolumn{1}{c|}{\multirow{2}{*}{AP}}& \multirow{2}{*}{Overall} & \multicolumn{4}{c|}{$\# Objects < 4$} & \multicolumn{4}{c|}{$4 \leq \# Objects < 7$} & \multicolumn{4}{c|}{$\# Objects \geq 7$} & \multirow{2}{*}{Short} & \multirow{2}{*}{Mid} & \multirow{2}{*}{Long} \\
& & Overall & Short & Mid & Long & Overall & Short & Mid & Long & Overall & Short & Mid & Long & & & \\ \hline

bike & 53.97 & 79.17 & 50.00 & 66.67 & 100.0 & 43.79 & 50.00 & 36.28 & 6.67 & 43.80 & N/A & 31.82 & N/A & 50.00 & 42.39 & 65.08 \\ 
cone & 19.49 & 36.36 & 50.0 & 16.67 & 0.00 & 47.60 & 55.56 & 60.00 & N/A & 2.12 & 0.00 & 3.81 & N/A & 23.71 & 18.16 & 0.00 \\ 
dog & 34.32 & 53.36 & 77.78 & 77.35 & 12.12 & 31.09 & 50.00 & 37.01 & 0.00 & 18.18 & 60.0 & N/A & 0.00 & 64.00 & 47.33 & 3.33 \\ 
mannequin & 36.91 & 70.57 & 64.00 & 85.71 & 16.67 & 21.51 & 32.73 & 26.67 & 5.83 & 39.66 & 0.00 & 52.94 & 29.41 & 29.57 & 48.72 & 14.22 \\ 
sign & 81.84 & 81.86 & 0.00 & 89.47 & 66.67 & 81.65 & N/A & 84.88 & 77.17 & 83.89 & N/A & 83.33 & 84.5 & 0.00 & 85.02 & 79.31 \\ 
trolley & 75.55 & 77.30 & 67.42 & 87.72 & 71.56 & 81.56 & 97.44 & 87.32 & 56.73 & 71.33 & 79.56 & 74.49 & 13.33 & 82.46 & 80.78 & 51.62 \\ 
\hline 
mAP & 50.35 & 66.44 & 51.53 & 70.60 & 44.50 & 51.20 & 57.14 & 55.36 & 24.40 & 43.16 & 34.89 & 49.28 & 25.45 & 41.62 & 53.73 & 35.60 \\ 

\end{tabular}}
\end{table*}

\begin{table*}[ht!]
\caption{CFAR+DBSCAN Detector Hard}\label{tab:dbscan_hard}
\resizebox{\textwidth}{!}{%
\begin{tabular}{l|c|c:ccc|c:ccc|c:ccc|c|c|c}
\multicolumn{1}{c|}{\multirow{2}{*}{AP}}& \multirow{2}{*}{Overall} & \multicolumn{4}{c|}{$\# Objects < 4$} & \multicolumn{4}{c|}{$4 \leq \# Objects < 7$} & \multicolumn{4}{c|}{$\# Objects \geq 7$} & \multirow{2}{*}{Short} & \multirow{2}{*}{Mid} & \multirow{2}{*}{Long} \\
& & Overall & Short & Mid & Long & Overall & Short & Mid & Long & Overall & Short & Mid & Long & & & \\ \hline

bike & 37.71 & 52.77 & 16.67 & 62.96 & 50.00 & 51.29 & 20.00 & 44.48 & 30.00 & 16.78 & N/A & 7.11 & N/A & 14.41 & 36.11 & 27.76 \\ 
cone & 6.35 & 8.33 & 25.00 & 0.00 & 0.00 & 17.65 & 22.22 & 16.67 & N/A & 0.00 & 0.00 & 0.00 & N/A & 10.34 & 3.7 & 0.00 \\ 
dog & 31.00 & 43.49 & 58.73 & 42.86 & 0.00 & 30.77 & 50.00 & 16.67 & 0.00 & 9.09 & 10.00 & N/A& 0.0 & 37.09 & 35.0 & 0.0 \\ 
mannequin & 7.95 & 37.39 & 66.67 & 35.71 & 0.00 & 3.92 & 0.00 & 6.67 & 0.00 & 0.00 & 0.00 & 0.00 & 0.00 & 16.67 & 8.97 & 0.0 \\ 
sign & 62.27 & 75.73 & 100.00 & 79.31 & 66.67 & 63.91 & N/A & 68.79 & 61.05 & 53.46 & N/A & 41.85 & 75.07 & 100.00 & 62.68 & 64.77 \\ 
trolley & 65.8 & 64.52 & 90.48 & 80.7 & 39.53 & 63.46 & 97.44 & 90.54 & 20.33 & 70.55 & 81.15 & 76.55 & 9.76 & 86.18 & 78.98 & 20.14 \\ 
\hline 
mAP & 35.18 & 47.04 & 59.59 & 50.26 & 26.03 & 38.50 & 37.93 & 40.63 & 18.56 & 24.98 & 22.79 & 25.10 & 16.97 & 44.12 & 37.58 & 18.78 \\ 

\end{tabular}}
\end{table*}


As expected, the results from a scene containing many known objects and confusing artefacts are much poorer than when the objects are classified from images of isolated objects.
Nevertheless, the results show promise.  For example, considering the mid range, \textit{Perfect Detector} case, there is an overall mean average precision of $61.36\%$,
and for specific easily distinguishable objects such as the trolley it is as high as $97.06\%$ in one instance. Other objects are more confusing, for example cones usually have low return power and can be easily confused with other small objects. As also expected the results degrade at long range and in scenes with a higher density of objects.

The \textit{Easy} case shows performance comparable but not as good as the \textit{Perfect Detector}, for example the mean average precision dropping to $50.35\%$. The CFAR + DBSCAN method is a standard option to detect objects in radar, but it does introduce some mistakes where, for example, the bounding box is misplaced with respect to the learnt radar patterns. 

Regarding the \textit{Hard} case, the mAP drops significantly to $35.18\%$. This shows how hard it is to recognise objects in radar images when the scene contains other, unseen and un-learnt, objects. Indeed, when the density of objects is greater than 7, some mAP values for bike, cone and mannequin are actually 0.00, which means that those objects were not recognised under those specific conditions.

Finally, we observe that trolley is the easiest object to recognise in all case. The trolley has a very characteristic shape, and strongly reflecting metal corner sections that create a distinguishable signature from all other objects. In interpreting true and false results in non-standardised datasets, which is the case in radar as opposed to visible camera imagery, one should be careful when comparing diverse published material.

\section{Conclusions}

In this work we evaluated the use of DCNNs applied to images from a 300 GHz radar system to recognise objects in a laboratory setting. Four types of experiments were performed to assess the robustness of the network.
These included the optimal scenario when all data is available for training and testing at different ranges, different viewing angles, and using different receivers. As expected, this performs best when all the training and test data are drawn from the same set.
This is a valuable experiment as it sets an optimal benchmark, but this is not a likely scenario for any radar system applied in the wild, first because radar data is far less ubiquitous or consistent than camera data, and second because the influence of clutter (really semantic background) and multipath effects are potentially more serious than for optical technology.

Regarding the single object scene data, we should be encouraged by two principal results, first that the performance was so high for the optimal case, and second that transfer learning may lead to improvements in other cases,
Transfer learning can prevent overfitting to the 300 GHz source data, by generalizing using more samples from a different radar data set, e.g.
increasing from $92.5\%$ to $98.5\%$ in the experiment using Q1 and Q3 to train and Q2 and Q4 to test. This leads to more robust classification. 

The multiple object dataset is a very challenging scenario, but we achieved mean average precision rates in the easy case $> 60\% (< 4 objects)$, but much less, $35.18\%$, in a high cluttered scenario. However, the pipeline we have adopted is probably subject to improvement, in particular is using the classification results to feed back to the detection and clustering. To avoid problems with occlusion, object adjacency, and multi-path, further research on high resolution radar images is necessary.
We also note that we have not made use of Doppler processing, as this implies motion of the scene, the sensor or both. For automotive radar, there are many stationary objects (e.g a car at a traffic light), and many different motion trajectories in the same scene, so this too requires further research.

In conclusion, it is very challenging in radar imagery for the deep learning approach to learn features which are robust to sensor height, type, range and orientation.
In the wild, by which we mean outside the laboratory and as a vehicle mounted sensor navigating the road network, we anticipate even more problems due to overall object density and proximity of targets to other scene objects.

\section*{Acknowledgements}

We acknowledge the support of the Engineering and Physical Research Council, grant reference EP/N012402/1 (TASCC: Pervasive low-TeraHz and Video Sensing for Car Autonomy and Driver Assistance (PATH CAD)).
We acknowledge particularly the work done by the Birmingham group led by Marina Gashinova in designing and building the 300GHz radar used for these experiments, and the practical help of Liam Daniel and Dominic Phippen in operating the radar and advising us on our experimental design. We also would like to thank Dr. David Wright for the phase correction code. Thanks too to NVIDIA for the donation of the TITAN X GPU.

\bibliographystyle{unsrt}
\bibliography{egbib}

\end{document}